\documentclass{article}

\usepackage[final]{neurips_2026}
\makeatletter
\renewcommand{\@noticestring}{Preprint.}
\makeatother

\usepackage[utf8]{inputenc}
\usepackage[T1]{fontenc}
\usepackage{amsmath,amssymb}
\usepackage{microtype}
\usepackage{graphicx}
\usepackage{booktabs}
\usepackage{array}
\usepackage{tabularx}
\usepackage{xcolor}
\usepackage{url}
\usepackage{caption}
\usepackage{enumitem}

\newcolumntype{L}{>{\raggedright\arraybackslash}X}

\usepackage{tikz}
\usetikzlibrary{arrows.meta,positioning,fit,calc,shapes}

\usepackage[colorlinks=true,linkcolor=blue,citecolor=blue,urlcolor=blue]{hyperref}

\title{An Architecture for Long-Horizon Agents:\\Levels, Ticks and Cascaded Intelligence}

\author{%
  Erik Nijkamp \\
  Salesforce AI Research \\
  \texttt{erik.nijkamp@salesforce.com}
  \And
  Anurag Koul \\
  Salesforce AI Research \\
  \texttt{anurag.koul@salesforce.com}
  \And
  Egor Pakhomov \\
  Salesforce AI Research \\
  \texttt{epakhomov@salesforce.com}
  \And
  Bo Pang \\
  Salesforce AI Research \\
  \texttt{b.pang@salesforce.com}
}

\begin{document}
\maketitle

\begin{abstract}
Language-model agents are increasingly asked to carry out work spanning days
or weeks, such as an operations remediation or a research programme. Such a
task outlives any context window, any process and any interval at which a
person can attend. In this paper, we argue that a long-horizon agent must run
continually without forgetting before it can learn continually. This ability
lies in the harness around the model rather than in the model itself. We
derive seven bottlenecks from the long-horizon setting and answer them with a
hierarchical architecture of three parts: (i) levels indexed by time scale,
each keeping a bounded file summarising the level below; (ii) a clocked tick
as the unit of autonomous action; and (iii) cascaded intelligence, where work
is escalated to a more capable model only after failing review. We report on
a ten-day campaign in which an agent built on this architecture reproduced a
published reinforcement-learning result with a human attending once a day,
and show (1) the agent kept the thread across every context reset and session
boundary of the campaign, (2) operating knowledge written early changed later
behaviour with no change to model weights, and (3) where learned components
would enter such a system. Overall, our experience suggests continual learning for these
agents needs a substrate outliving every context and process, and the checks
the harness already runs are where a learner belongs.
\end{abstract}

\section{Introduction: Agents Beyond the Episode}
\label{sec:intro}

Language-model agents now carry out multi-step work in software engineering,
research and operations \citep{react2022,openhands}, and the horizon they
can sustain has grown from minutes to hours in two years
\citep{metrhorizon,rebench}. Deployed agents nonetheless act within an episode: one context window, then
a stop, with initiative, memory between episodes and the next start left to
the person who asked. Enterprise work does not have this shape: an
operations remediation outlives the shift that opened it, a compliance
investigation runs for a week, a research programme for months. Longer
contexts and better models push the episode boundary outward but do not remove it: compaction
erases state and constraints silently \citep{governancedecay}, agents drift
from the goal as the horizon grows \citep{goaldrift,horizonmirage},
and however long the context, the campaign, the task as a whole, is
longer.

Prior work answers parts of this problem. Memory architectures decide what an
agent remembers across sessions \citep{memgpt2023,generativeagents2023,memorysurvey};
hierarchical agents decompose a long task into sub-goals
\citep{options1999,himac}; model cascades route work to a stronger model by
the cost of being wrong \citep{humancascade,escalationworthit}; and
autonomous research agents run whole campaigns end to end
\citep{aiscientist2024,mlreplicate}. Each lives inside one context or process; durable execution
\citep{crashonly} outlasts the process but has no notion of what to keep. None says what must survive when context,
process and attention each end; that gap is architectural.

In this paper, we take the position that continual learning for long-horizon
agents rests on a prior property: continual operation without forgetting. This
property belongs to the harness, the system around the model that decides when it runs, what it reads and what it may do. Studies of agent
failures \citep{mast,harnessfail} trace much of what goes wrong to this layer,
not the model. From one premise, that an agent outlives any context, process or interval of
human attention, we derive seven bottlenecks and present a hierarchical
architecture answering them. An agent built on this
architecture ran for ten days and reproduced a published result
\citep{sao2026}; it updated no weight, yet what it wrote down early changed
what it did later. We make three contributions.
\begin{enumerate}[leftmargin=*,nosep]
\item \textbf{Characterisation} --- seven bottlenecks that follow from the
long-horizon premise (\S\ref{sec:regime}).
\item \textbf{Architecture} --- four abstractions (level, tick, protocol,
tier) that answer the bottlenecks and yield operation without forgetting,
accumulation, and the points where a learner enters (\S\ref{sec:architecture}, \S\ref{sec:position}).
\item \textbf{Experience} --- ten days in which an agent reproduced a published
reinforcement-learning result on its own, over two hundred ticks and about
two dozen escalations to the human (\S\ref{sec:experience}).
\end{enumerate}


\section{Setting: From One Premise to Seven Bottlenecks}
\label{sec:regime}

\begin{figure}[t]
\centering
\resizebox{0.84\textwidth}{!}{
\begin{tikzpicture}[
    font=\small,
    >={Stealth[length=2.2mm]},
    ax/.style={font=\footnotesize,text=black!60},
    thr/.style={black!40,densely dashed,line width=0.5pt},
    hd/.style={font=\footnotesize,text=black!65,anchor=south,align=center},
    kw/.style={font=\small\bfseries,text=black!85,align=center},
    mech/.style={font=\footnotesize,text=black!65,align=center},
    hum/.style={font=\footnotesize,text=black!60,align=center},
    reg/.style={font=\small,text=black!70,align=center},
    push/.style={->,black!60,line width=0.8pt}]
  \def\yb{3.0}   
  \def\yt{6.0}   
  \def\xr{14.2}  
  \fill[black!4,rounded corners=3pt] (2.7,\yb+0.08) rectangle (\xr-0.1,\yt-0.05);
  \node[reg,text=black!75] at (10.7,5.35) {autonomous long-horizon agent};
  \draw[->,black!60,line width=0.6pt] (0,0) -- (\xr+0.4,0) node[anchor=north east,ax,yshift=-2pt] {horizon $H$: length of the trajectory (actions, tokens, or wall time)};
  \draw[->,black!60,line width=0.6pt] (0,0) -- (0,\yt+1.45);
  \node[ax,rotate=90,anchor=south] at (-1.9,3.3) {autonomy: who supplies initiative, delegation and checking};
  \node[ax,anchor=east] at (-0.15,1.5) {the human};
  \node[ax,anchor=east] at (-0.15,4.5) {the agent};
  \foreach \x/\lab in {2.6/{context\\window $C$},4.9/{process\\lifetime $P$},7.2/{human attention\\interval $A$},9.5/{failure\\interval $F$},11.8/{budget $B$ at\\top-tier cost}}{
    \draw[thr] (\x,0) -- (\x,\yt+0.55);
    \node[hd] at (\x,\yt+0.6) {\lab};
  }
  \draw[black!70,line width=0.9pt] (0,\yb) -- (\xr,\yb);
  \node[font=\footnotesize,text=black!80,fill=white,inner sep=2pt] at (7.1,\yb) {\textbf{Autonomy}: initiative moves from the human to the agent};
  \foreach \x/\k/\m in {3.75/{Horizon}/{abstraction\\hierarchy of levels},6.05/{State}/{files as memory,\\resume},8.35/{Delegation}/{briefs, steering,\\escalation},10.65/{Resilience}/{watchdogs,\\inert probes},13.0/{Cost}/{cascaded\\intelligence}}{
    \node[kw] at (\x,4.4) {\k};
    \node[mech] at (\x,3.7) {\m};
  }
  \node[reg] at (1.3,4.1) {agentic\\episode};
  \foreach \x/\m in {3.75/{re-prompts,\\summarises},6.05/{restarts,\\re-explains},8.35/{orchestrates,\\decides},10.65/{notices,\\repairs},13.0/{pays,\\or stops}}{
    \node[hum] at (\x,0.75) {\m};
  }
  \node[reg] at (1.3,1.5) {episodic\\assistant};
  \node[reg] at (8.35,2.1) {human-driven campaign};
  \draw[black!60,line width=0.6pt] (2.7,\yt+0.15) -- (\xr-0.1,\yt+0.15);
  \draw[black!60,line width=0.6pt] (2.7,\yt+0.02) -- (2.7,\yt+0.15);
  \draw[black!60,line width=0.6pt] (\xr-0.1,\yt+0.02) -- (\xr-0.1,\yt+0.15);
  \node[font=\footnotesize,text=black!80,fill=white,inner sep=2pt] at (8.4,\yt+0.15) {\textbf{Correctness}: no one checks a step, errors compound, compression hides them};
  \draw[push] (2.6,\yt+1.75) -- (\xr-0.1,\yt+1.75) node[midway,ax,above=2pt] {push the horizon: one resource after another runs out};
  \draw[push] (\xr+0.55,0.3) -- (\xr+0.55,\yt-0.2) node[midway,ax,rotate=90,above=2pt] {push autonomy: the human stops supplying};
\end{tikzpicture}

}
\caption{The long-horizon setting. Each threshold along the horizon is where a
resource runs out; along autonomy, the question is who supplies the function when it does. Below the
line the person does; above it the architecture must, and each column names the bottleneck it must answer.}
\label{fig:paradigm}
\end{figure}
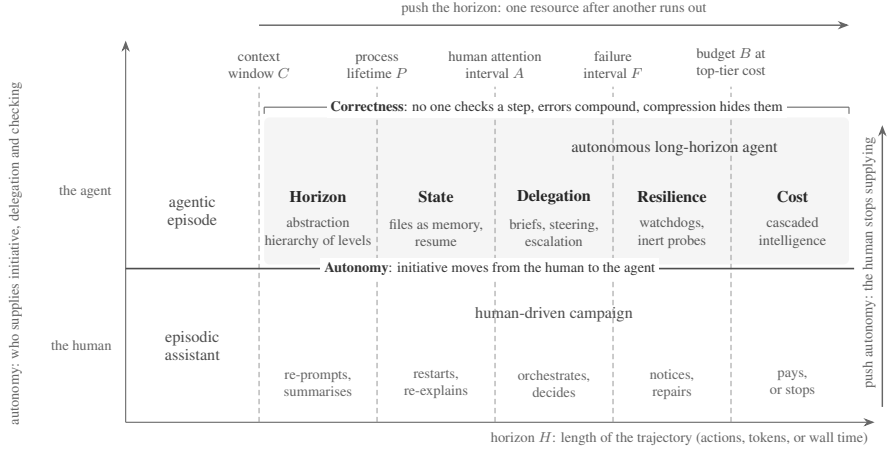

In this section we place agents on two axes, autonomy and horizon
(Fig.~\ref{fig:paradigm}), and derive the seven bottlenecks that arise when
both are high. Each becomes a requirement on the architecture of
\S\ref{sec:architecture}.

\paragraph{Two axes.} \emph{Autonomy} is the share of decisions an agent takes without a person: at
its high end an \emph{active} agent holds a goal, decides when and how to act and continues until the goal is
met or a constraint stops it, while a \emph{passive} principal sets the goal,
reads what is reported and answers what is escalated. The \emph{horizon} $H$
is the length of the trajectory the agent must sustain to reach the goal,
measured in actions, in tokens, or in wall time at a given action rate.

\paragraph{Thresholds.} What makes this setting different is not the size of
$H$ but what the trajectory crosses on the way. Four quantities lie on the horizon axis: the context window $C$ of one model call, the
lifetime $P$ of one process or session, the interval $A$ at which the
principal attends, and the failure interval $F$ of any component the agent
depends on. Each has an enterprise reading: a ticket that outlives the conversation that
opened it, a job that outlives its container, a decision that waits for a
manager, a system that fails mid-campaign. At high autonomy no one absorbs a crossing, and each threshold becomes a
requirement on the architecture.

\paragraph{Correctness and cost.} Two further bottlenecks grow with both axes
at once. \emph{Correctness}: autonomy removes the person who would check a
step, and the horizon compounds whatever goes unchecked while compression
removes the evidence from view. \emph{Cost}: spend at the most capable tier
grows with $H$ at a rate $c_{\mathrm{top}}$ per unit of horizon and the budget
$B$ caps it, so the horizon affordable at top-tier cost is
$B/c_{\mathrm{top}}$, and cascading work to cheaper tiers moves that ceiling
outward.

\paragraph{Seven bottlenecks.} The setting we study is $C, P, A, F \ll H$ at
high autonomy within $B$, and seven bottlenecks follow: four from the
thresholds, Autonomy ($A \ll H$), Horizon ($C \ll H$), State ($P \ll H$) and
Resilience ($F \ll H$); one from scale, Delegation (the work exceeds one
process); and Correctness and Cost from both axes.
Table~\ref{tab:requirements} in the appendix derives each requirement.


\section{Architecture: Levels, Ticks, Protocols and Tiers}
\label{sec:architecture}

\subsection{Four Abstractions and Their Relations}
\label{sec:abstractions}
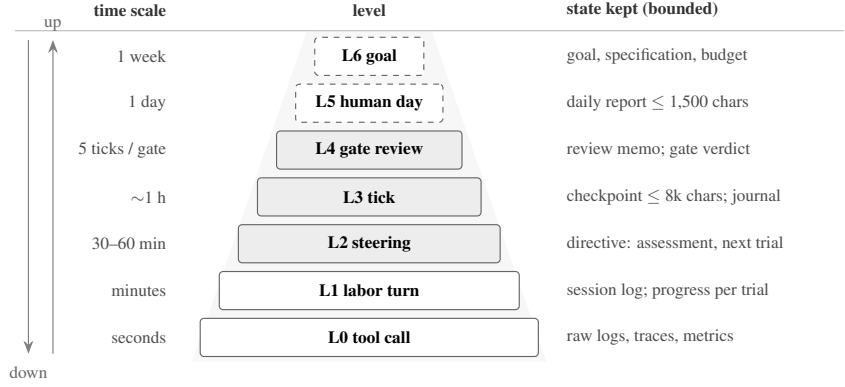
\begin{figure}[t]
\centering
\resizebox{0.8\textwidth}{!}{
\begin{tikzpicture}[
    font=\small,
    >={Stealth[length=2.4mm]},
    lvl/.style={draw=black!60,line width=0.6pt,rounded corners=2pt,align=center,
                minimum height=7mm,inner xsep=4pt},
    hum/.style={lvl,fill=white,dashed},
    jud/.style={lvl,fill=black!7},
    lab/.style={lvl,fill=white},
    prc/.style={lvl,fill=white},
    ts/.style={anchor=east,font=\small,text=black!70},
    st/.style={anchor=west,font=\small,text=black!70,align=left},
    hd/.style={font=\small\bfseries,text=black!80},
    ch/.style={line width=0.6pt,black!50}]

  \def\rs{0.86}
  \fill[black!3] (-3.25,-0.45) -- (3.25,-0.45) -- (1.15,6*\rs+0.45) -- (-1.15,6*\rs+0.45) -- cycle;

  \foreach \i/\sty/\name/\w/\tsc/\state in {
     0/prc/{L0 tool call}/6.2/{seconds}/{raw logs, traces, metrics},
     1/lab/{L1 labor turn}/5.5/{minutes}/{session log; progress per trial},
     2/jud/{L2 steering}/4.8/{30--60 min}/{directive: assessment, next trial},
     3/jud/{L3 tick}/4.1/{$\sim$1 h}/{checkpoint $\le$ 8k chars; journal},
     4/jud/{L4 gate review}/3.4/{5 ticks / gate}/{review memo; gate verdict},
     5/hum/{L5 human day}/2.7/{1 day}/{daily report $\le$ 1{,}500 chars},
     6/hum/{L6 goal}/2.0/{1 week}/{goal, specification, budget}}
  {
     \node[\sty,minimum width=\w cm] (r\i) at (0,\i*\rs) {\textbf{\name}};
     \node[ts] at (-3.6,\i*\rs) {\tsc};
     \node[st] at (3.5,\i*\rs) {\state};
  }

  \def\hy{6*\rs+0.85}
  \node[hd,anchor=east] at (-3.6,\hy) {time scale};
  \node[hd] at (0,\hy) {level};
  \node[hd,anchor=west] at (3.5,\hy) {state kept (bounded)};
  \draw[black!30] (-6.5,\hy-0.38) -- (8.7,\hy-0.38);

  \draw[->,ch] (-6.25,6*\rs+0.3) -- (-6.25,-0.3);
  \draw[->,ch] (-5.8,-0.3) -- (-5.8,6*\rs+0.3);
  \node[font=\small,text=black!60,anchor=north] at (-6.25,-0.4) {down};
  \node[font=\small,text=black!60,anchor=south] at (-5.8,6*\rs+0.35) {up};
\end{tikzpicture}

}
\caption{The temporal hierarchy: seven levels, from a tool call at seconds
to the goal at a week. Each keeps a bounded file that compresses the level
below. Directives travel down; summaries, escalations and attention items travel up. Workers and processes act at L0--L1, the judgment tier at L2--L4,
the principal at L5--L6.}
\label{fig:timescales}
\end{figure}

The architecture rests on four abstractions. A \emph{level} is a layer of the
system indexed by the time scale at which it acts; it keeps a bounded state
that summarises the level below and issues sparse directives to it. The
\emph{tick} is the unit of autonomous action at the level where judgment
resides: one wake of the \emph{driver}, the judgment-tier session that runs
the loop, from reading state to writing it back. A
\emph{protocol} is a named file with a designated writer, reader, cadence and
check; every channel and every piece of durable state is a protocol. There are three \emph{tiers}
of intelligence: judgment, labor (routine or strong; its sessions are
\emph{workers}) and deterministic processes. Tier rises with level. Fig.~\ref{fig:timescales} shows the levels that result and the file each
keeps.

The seven bottlenecks name seven properties of one object, the level: a time
scale and a compression ratio (Horizon), a bounded file (State), an actor of a
given tier (Cost), channels to its neighbours (Delegation), a trigger
(Autonomy), a verification of the level below (Correctness), and a failure
mode with a detector one level up (Resilience). Downward channels carry
\emph{steering}, sparse directives at the sender's cadence; upward channels
carry \emph{back-pressure}, compressed state plus escalation lines, attention
items raised by monitors, and kill verdicts, the only messages that cross a boundary uncompressed,
because they are the ones compression would hide. The tick is where all seven meet.

\subsection{Seven Mechanism Families}
\label{sec:mechanisms}

\paragraph{Autonomy.} Initiative must be manufactured. A clock starts a fresh driver session and sleeps for the interval the driver
wrote before exiting; a tick reads state, applies every pre-registered rule,
decides, acts and writes state back, and a hook refuses to end the session
before the tick-level state file, the checkpoint, is written. The principal's answers are kept as
\emph{standing decisions}, read at every wake, so the agent proceeds through the principal's absence.

\paragraph{Horizon.} A horizon of days is handled by abstraction indexed by
time scale: each level sees a bounded compression of the level below and acts
at its own cadence; the driver never reads raw output. The rule is hard:
at most forty lines of any log or source; the rest is delegated and
summarised.

\paragraph{State.} Nothing may live only in a context. Each level's state is a file, bounded so that the level above can afford to read it, rewritten in
place so that it describes the present, with every number naming its source file. A resume protocol
runs at every session start and after every compaction of the driver's
context: checkpoint first, then standing decisions, plan, open unknowns and
active experiment cards; if the append-only journal is newer than the
checkpoint, the journal wins.

\paragraph{Delegation.} Work travels down as a contract and returns as
evidence; decisions travel up, ultimately to the principal. A \emph{brief} is one file: task, goal, tier, files to read,
constraints, what to return. For open-ended work \emph{steered labor} adds two files: the worker writes
\textsc{progress}, plan before and measured result after each trial; the
driver writes \textsc{directive} at the steering cadence, which hypothesis the
numbers support and what to try next, re-read by the worker before every
action. Decisions climb a ladder with a stated trigger per rung: routine
labor, the driver's own assessment, a strong-tier session, the principal, for a listed class of decisions only.

\paragraph{Correctness.} The agent tunes blind unless
verification is pre-registered, measured and placed where its result cannot be
lost. Every experiment's card carries a hypothesis, success criteria, an early-stop
rule, a resource cap and a seed, and the driver applies the
rule at every tick, so a kill verdict is a bit in a summary. A run counts as comparable only when its \emph{regime conditions} are
measured rather than requested, such as realised staleness. An \emph{adversarial reviewer}, a judgment-tier session with no
conversation history, reads plan and evidence at every \emph{gate}, a
pre-registered check before compute or before a claim: which confounds are open, whether the proxy measures the claim, what to
kill.

\paragraph{Resilience.} Every component fails at least once over days,
including the agent's own sessions, so recovery cannot depend on attention. Each level has a detector one level up and a judgment-free recovery:
a stalled worker is re-briefed, a dead driver resumes from the checkpoint
under a watchdog that halts after repeated restarts, and probes run as steps
of the job they observe, never as logins.

\begin{figure}[t]
\centering
\resizebox{0.52\textwidth}{!}{
\begin{tikzpicture}[
    font=\normalsize,
    >={Stealth[length=2.2mm]},
    box/.style={draw=black!60,line width=0.6pt,rounded corners=2pt,fill=white,align=center,minimum height=9mm,minimum width=23mm},
    dec/.style={draw=black!60,line width=0.6pt,fill=black!5,align=center,minimum height=9mm,minimum width=20mm},
    hum/.style={box,dashed},
    ch/.style={->,line width=0.6pt,black!55},
    lab/.style={font=\small,text=black!60,fill=white,inner sep=1.5pt,align=center}]
  \node[box] (B) at (0,0) {brief\\{\small\color{black!60} tier per brief}};
  \node[box] (R) at (3.1,0) {routine tier\\{\small\color{black!60} executes}};
  \node[dec] (V1) at (6.2,0) {review\\{\small\color{black!60} one tier up}};
  \node[box,fill=black!5] (OK) at (9.3,0) {accepted\\{\small\color{black!60} result, commit}};
  \node[box] (S) at (6.2,-2.0) {strong tier\\{\small\color{black!60} diagnoses, retries}};
  \node[dec] (V2) at (9.3,-2.0) {review};
  \node[hum] (H) at (9.3,-4.0) {principal\\{\small\color{black!60} listed decisions}};
  \draw[ch] (B) -- (R);
  \draw[ch] (R) -- (V1);
  \draw[ch] (V1) -- (OK) node[midway,lab,above=1pt]{pass};
  \draw[ch] (V1.north) to[bend right=35] node[pos=0.5,lab,above=1pt]{fail once: feedback, retry} (R.north);
  \draw[ch] (V1) -- (S) node[midway,lab,right=1pt]{fail again: raise a tier};
  \draw[ch] (S) -- (V2);
  \draw[ch] (V2) -- (OK) node[midway,lab,right=1pt]{pass};
  \draw[ch] (V2.south west) to[bend left=35] node[pos=0.5,lab,below=1pt]{feedback, retry} (S.south east);
  \draw[ch] (V2) -- (H) node[pos=0.72,lab,left=1pt]{fail again, or not the agent's call};
\end{tikzpicture}

}
\caption{Cascaded intelligence with review-and-escalate: work runs at the
lowest plausible tier and is reviewed one tier up; failure returns as feedback
for a retry, a second failure raises it one tier, up to the principal.}
\label{fig:cascade}
\end{figure}
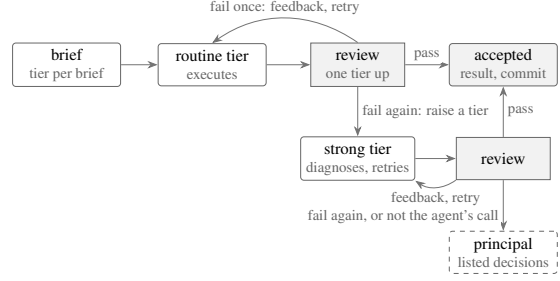

\paragraph{Cost.} The most capable model attends only to the most compressed
state; the rule that moves work between tiers is \emph{review-and-escalate} (Fig.~\ref{fig:cascade}): work is done at
the lowest tier that can plausibly do it, reviewed one tier up, returned with
feedback if it fails, and raised one tier only if it fails again. The strong
tier is assigned by the cost of being wrong rather than the size of the task,
and the ladder is symmetric: once the strong tier has found a recipe,
successors run at the routine tier with the recipe in their brief. Sessions are rotated every few ticks, so a fresh session reads a short checkpoint, and a deterministic
pre-tick digest skips a wake when nothing is due \citep[cf.][]{whentowake}.

\subsection{Operation Without Forgetting, and Accumulation}
\label{sec:continual}
The agent does not learn in the parametric sense; two weaker properties
hold, and both follow from the hierarchy. The first is \emph{no forgetting}. Compaction erases constraints silently \citep{governancedecay,compactiontheory};
here nothing the agent needs lives in the context. Every level keeps its bounded compression in a file and every number names
its source; the driver was reset dozens of times, by compaction, rotation or
watchdog, and resumed from files each time. The
second is \emph{accumulation}: operating knowledge written at one level
changes behaviour at the levels below it for the rest of the campaign. The principal's answers become standing decisions; reviewer findings become
pre-registered rules on later cards; a recipe found by the strong tier is
handed down into routine briefs; a recurring worker failure becomes a rule in
every brief and then a guard in the harness (\S\ref{sec:experience}). This is accumulation, not learning, but it fixes where a learned component
would sit: inside a gate that stays (\S\ref{sec:position}).


\section{Experience: Ten Autonomous Days on One Hard Task}
\label{sec:experience}

\paragraph{The task.} The agent was asked to
reproduce a published result on asynchronous reinforcement learning for
agentic tasks \citep{sao2026}: that a policy trained on stale rollouts, sampled from an older
policy, collapses beyond a staleness threshold and that a decoupled
importance-sampling correction keeps it stable, at two model scales. The task has the properties that make an enterprise campaign hard: a
verifiable outcome, shared infrastructure, a fixed budget, a principal who
attends once a day, and failure modes that do not announce themselves.

\paragraph{The campaign.} Over ten days (Table~\ref{tab:campaign} in the appendix) the campaign moved
through five gated phases: bring-up of the environment;
the stability pair, uncorrected against corrected, at the small scale across
seeds; a one-node recipe for the large model, found under steered labor, and
the stability pair at that scale; a port to a second framework, validated by
realised rather than requested staleness and optimised to match the run it
reproduced on half the training hardware; and the effectiveness
comparison, full method against control, on a calibrated environment that went through four redesigns under review gates. The outcome reproduced the published result: the uncorrected
policy collapsed beyond the staleness threshold at both scales and the
corrected one held, and the full method reached the ceiling of the calibrated environment ahead of
its control.

\paragraph{Three vignettes.}
\emph{Steered labor.} The large model had to train on one node, and three
unsteered worker sessions had ended at 958 seconds per step with no
diagnosis. The driver wrote six ranked hypotheses, each with a prediction and a guard,
then a directive every thirty minutes while one worker ran trials. Within one bounded session the step time fell to 180 seconds
and the recipe held without the strong tier.
\emph{The reviewer.} When the effectiveness comparison saturated, the driver
commissioned a harder task mix and the reviewer read it first: it found a path by which the verifier could be satisfied without solving the task
and two families a model could shortcut; the mix was rebuilt. Later it found that a planned stop rule would have compared an arm against
its control's transient dip, and the rule was rewritten before the result was recorded.
\emph{Accumulation.} A worker that watches a running job must not end its turn
before the job does, or a node idles until the next tick. The first time, the
rule went into every worker brief; the second time, it became a
harness stop guard that refuses to end a worker owning a live run, and the
failure did not recur.


\section{Position: A Substrate Before a Learner}
\label{sec:position}
Our position is that before a long-horizon agent can learn continually it
must run continually without forgetting, and that this is a property of the
harness. The campaign in \S\ref{sec:experience} is the evidence that such a
harness can be built. When the campaign moved to a second training framework, only launch scripts
and an adapter changed; every brief, card, directive and review format carried
over. The same files that kept the agent from forgetting kept it inside what it was
told: an authenticated principal and a class of decisions it may not take. The limits are those of a single deployment: one campaign, an existence proof
rather than a benchmark.

Three consequences follow for continual learning in long-horizon agents.
First, \emph{the record is the dataset}: every tick leaves a checkpoint, every
brief a review outcome, every card a verdict. Second, \emph{the gates are the
insertion points}: a learned policy for the tier of a brief or the interval of
a wake replaces a written rule inside a check that stays, so what it learns
cannot outrun what the harness verifies. Third, \emph{accumulation without
learning is the baseline}: a method that updates weights must keep more of a
campaign across resets than a hierarchy that writes things down.


\bibliographystyle{plainnat}

\begin{thebibliography}{22}
\providecommand{\natexlab}[1]{#1}
\providecommand{\url}[1]{\texttt{#1}}
\expandafter\ifx\csname urlstyle\endcsname\relax
  \providecommand{\doi}[1]{doi: #1}\else
  \providecommand{\doi}{doi: \begingroup \urlstyle{rm}\Url}\fi

\bibitem[Arike et~al.(2025)Arike, Donoway, Bartsch, et~al.]{goaldrift}
Rauno Arike, Elizabeth Donoway, Henning Bartsch, et~al.
\newblock Technical report: Evaluating goal drift in language model agents.
\newblock \emph{arXiv preprint arXiv:2505.02709}, 2025.

\bibitem[Bouchard(2026)]{escalationworthit}
Dylan Bouchard.
\newblock Is escalation worth it? a decision-theoretic characterization of llm
  cascades.
\newblock \emph{arXiv preprint arXiv:2605.06350}, 2026.

\bibitem[Candea and Fox(2003)]{crashonly}
George Candea and Armando Fox.
\newblock Crash-only software.
\newblock In \emph{9th Workshop on Hot Topics in Operating Systems (HotOS IX)}.
  USENIX Association, 2003.
\newblock URL
  \url{https://www.usenix.org/legacy/events/hotos03/tech/full_papers/candea/candea.pdf}.

\bibitem[Cemri et~al.(2025)Cemri, Pan, Yang, Agrawal, Chopra, Tiwari, Keutzer,
  Parameswaran, Klein, Ramchandran, Zaharia, Gonzalez, and Stoica]{mast}
Mert Cemri, Melissa~Z. Pan, Shuyi Yang, Lakshya~A. Agrawal, Bhavya Chopra,
  Rishabh Tiwari, Kurt Keutzer, Aditya Parameswaran, Dan Klein, Kannan
  Ramchandran, Matei Zaharia, Joseph~E. Gonzalez, and Ion Stoica.
\newblock Why do multi-agent llm systems fail?
\newblock \emph{arXiv preprint arXiv:2503.13657}, 2025.
\newblock URL \url{https://arxiv.org/abs/2503.13657}.

\bibitem[Chen(2026)]{governancedecay}
Shiyang Chen.
\newblock Governance decay: How context compaction silently erases safety
  constraints in long-horizon {LLM} agents.
\newblock \emph{arXiv preprint arXiv:2606.22528}, 2026.

\bibitem[Fanconi and van~der Schaar(2025)]{humancascade}
Claudio Fanconi and Mihaela van~der Schaar.
\newblock Cascaded language models for cost-effective human-ai decision-making.
\newblock \emph{arXiv preprint arXiv:2506.11887}, 2025.

\bibitem[Gaddipati et~al.(2026)Gaddipati, Muhammed, Keya, Rabby, and
  Auer]{mlreplicate}
Sasi~Kiran Gaddipati, Diyana Muhammed, Farhana Keya, Gollam Rabby, and
  S{\"o}ren Auer.
\newblock Mlreplicate: Benchmarking autonomous research systems for machine
  learning reproducibility.
\newblock \emph{arXiv preprint arXiv:2605.16616}, 2026.

\bibitem[Hou et~al.(2026)Hou, Li, Tang, and Dong]{sao2026}
Zhenyu Hou, Yujiang Li, Jie Tang, and Yuxiao Dong.
\newblock Single-rollout asynchronous optimization for agentic reinforcement
  learning.
\newblock \emph{arXiv preprint arXiv:2607.07508}, 2026.

\bibitem[Hu et~al.(2025)Hu, Liu, Yue, Zhang, et~al.]{memorysurvey}
Yuyang Hu, Shichun Liu, Yanwei Yue, Guibin Zhang, et~al.
\newblock Memory in the age of {AI} agents.
\newblock \emph{arXiv preprint arXiv:2512.13564}, 2025.

\bibitem[Jin et~al.(2026)Jin, Zhu, Ding, Luo, and Li]{himac}
Hongbo Jin, Rongpeng Zhu, Jiayu Ding, Guibo Luo, and Ge~Li.
\newblock Himac: Hierarchical macro-micro learning for long-horizon llm agents.
\newblock \emph{arXiv preprint arXiv:2603.00977}, 2026.

\bibitem[Kwa et~al.(2025)Kwa, West, Becker, et~al.]{metrhorizon}
Thomas Kwa, Ben West, Joel Becker, et~al.
\newblock Measuring {AI} ability to complete long software tasks.
\newblock In \emph{Advances in Neural Information Processing Systems
  (NeurIPS)}, 2025.

\bibitem[Liu et~al.(2026)Liu, Zhang, Abdi, Galley, Chen, Xiong, Wang, and
  Gao]{whentowake}
Xiaoze Liu, Ruowang Zhang, Amir~H. Abdi, Michel Galley, Zhikai Chen, Siheng
  Xiong, Xiaoqian Wang, and Jing Gao.
\newblock Do proactive agents really need an {LLM} to decide when to wake and
  what to anchor?
\newblock \emph{arXiv preprint arXiv:2605.30152}, 2026.

\bibitem[Lu et~al.(2024)Lu, Lu, Lange, Foerster, Clune, and
  Ha]{aiscientist2024}
Chris Lu, Cong Lu, Robert~Tjarko Lange, Jakob Foerster, Jeff Clune, and David
  Ha.
\newblock The {AI} {Scientist}: Towards fully automated open-ended scientific
  discovery.
\newblock \emph{arXiv preprint arXiv:2408.06292}, 2024.

\bibitem[Packer et~al.(2023)Packer, Wooders, Lin, Fang, Patil, Stoica, and
  Gonzalez]{memgpt2023}
Charles Packer, Sarah Wooders, Kevin Lin, Vivian Fang, Shishir~G. Patil, Ion
  Stoica, and Joseph~E. Gonzalez.
\newblock {MemGPT}: Towards {LLMs} as operating systems.
\newblock \emph{arXiv preprint arXiv:2310.08560}, 2023.

\bibitem[Park et~al.(2023)Park, O'Brien, Cai, Morris, Liang, and
  Bernstein]{generativeagents2023}
Joon~Sung Park, Joseph~C. O'Brien, Carrie~J. Cai, Meredith~Ringel Morris, Percy
  Liang, and Michael~S. Bernstein.
\newblock Generative agents: Interactive simulacra of human behavior.
\newblock In \emph{Proceedings of the 36th Annual ACM Symposium on User
  Interface Software and Technology (UIST)}, 2023.
\newblock \doi{10.1145/3586183.3606763}.

\bibitem[Raj et~al.(2026)Raj, Gupta, Mahmoud, Dumitru, Yi, Sabharwal, and
  He]{harnessfail}
Harsh Raj, Vipul Gupta, Anas Mahmoud, Razvan-Gabriel Dumitru, Darvin Yi, Aakash
  Sabharwal, and Yunzhong He.
\newblock Model or harness? an interaction-centric taxonomy for localizing
  agent failures.
\newblock \emph{arXiv preprint arXiv:2607.28802}, 2026.
\newblock URL \url{https://arxiv.org/abs/2607.28802}.

\bibitem[Sutton et~al.(1999)Sutton, Precup, and Singh]{options1999}
Richard~S. Sutton, Doina Precup, and Satinder Singh.
\newblock Between mdps and semi-mdps: A framework for temporal abstraction in
  reinforcement learning.
\newblock \emph{Artificial Intelligence}, 112\penalty0 (1--2):\penalty0
  181--211, 1999.

\bibitem[Tirmazi et~al.(2026)Tirmazi, Markelon, Bishop, and
  Mitzenmacher]{compactiontheory}
Hayder Tirmazi, Sam Markelon, Allison Bishop, and Michael Mitzenmacher.
\newblock Context compaction theory.
\newblock \emph{arXiv preprint arXiv:2608.01326}, 2026.

\bibitem[Wang et~al.(2025)Wang, Li, Song, et~al.]{openhands}
Xingyao Wang, Boxuan Li, Yufan Song, et~al.
\newblock Openhands: An open platform for ai software developers as generalist
  agents.
\newblock In \emph{International Conference on Learning Representations
  (ICLR)}, 2025.

\bibitem[Wang et~al.(2026)Wang, Bai, Sun, et~al.]{horizonmirage}
Xinyu~Jessica Wang, Haoyue Bai, Yiyou Sun, et~al.
\newblock The long-horizon task mirage? diagnosing where and why agentic
  systems break.
\newblock \emph{arXiv preprint arXiv:2604.11978}, 2026.

\bibitem[Wijk et~al.(2024)Wijk, Lin, Becker, et~al.]{rebench}
Hjalmar Wijk, Tao Lin, Joel Becker, et~al.
\newblock {RE-Bench}: Evaluating frontier {AI} {R\&D} capabilities of language
  model agents against human experts.
\newblock \emph{arXiv preprint arXiv:2411.15114}, 2024.

\bibitem[Yao et~al.(2022)Yao, Zhao, Yu, Du, Shafran, Narasimhan, and
  Cao]{react2022}
Shunyu Yao, Jeffrey Zhao, Dian Yu, Nan Du, Izhak Shafran, Karthik Narasimhan,
  and Yuan Cao.
\newblock React: Synergizing reasoning and acting in language models.
\newblock \emph{arXiv preprint arXiv:2210.03629}, 2022.

\end{thebibliography}

\appendix
\section{The Campaign in Numbers, and the Seven Bottlenecks}
\label{app:bottlenecks}
Table~\ref{tab:campaign} tallies the completed campaign of
\S\ref{sec:experience}. Table~\ref{tab:requirements} derives each bottleneck of
\S\ref{sec:regime} from the premise and names the requirement it imposes and
the mechanism family that answers it.
\begin{table}[h]
\caption{The campaign in numbers: tallies from the project record over the
ten days.}
\label{tab:campaign}
\centering\footnotesize
\begin{tabular}{@{}lr@{\hspace{1.6em}}lr@{}}
\toprule
quantity & value & quantity & value \\
\midrule
calendar days / days under the clock & 10 / 8 & experiments & $\approx$ 40 \\
driver ticks / context resets & 211 / 47 & adversarial reviews / critical findings & 30 / 7 \\
worker sessions, routine / strong & 403 / 35 & escalations / principal-only decisions & 25 / 9 \\
analysis sessions at the judgment tier & 25 & GPU-hours used / cap & 2{,}215 / 3{,}000 \\
labor spend, routine / strong & 86\,\% / 14\,\% & & \\
\bottomrule
\end{tabular}
\end{table}


\begin{table}[h]
\caption{The seven bottlenecks of long-horizon autonomy. Each follows from the premise ($C, P, A, F \ll H$ at high autonomy within $B$), names a requirement, and is answered by a mechanism family in \S\ref{sec:mechanisms}.}
\label{tab:requirements}
\centering\footnotesize
\begin{tabular}{@{}lp{0.26\textwidth}p{0.27\textwidth}p{0.27\textwidth}@{}}
\toprule
concept & follows from & requirement & mechanism family \\
\midrule
Autonomy    & $A \ll H$: no one prompts & own clock, goal, policy, and continuation through absence & goal file, tick, wake scheduling, standing decisions \\
Horizon     & $C \ll H$: no context holds the trajectory & act on a compression organised by time scale & levels, bounded files, cadences, read limits \\
State       & $P \ll H$: every process ends first & nothing lives only in a context; rebuildable from files & memory hierarchy, resume protocol, snapshots, allocators \\
Delegation  & work $\gg$ one process; $A \ll H$: no human hands it on & contracts, steering while running, decisions upward & briefs, steered labor, back channels, escalation ladder, trust gate \\
Correctness & autonomy $\times$ horizon: no human checks a step; errors compound and their evidence is compressed away & verification that survives compression & pre-registration, kill rules, regime conditions, adversarial review \\
Resilience  & $F \ll H$: every component fails during the campaign & failure expected at every level, handled by protocol & wall times, watchdogs, monitors, inert probes, clean restarts \\
Cost        & autonomy $\times$ horizon: top-tier spend $\propto H$, capped by $B$ & cascaded intelligence with a rule for moving work & tiers by level, per-brief tier, review-and-escalate, cadence, caching \\
\bottomrule
\end{tabular}
\end{table}





\end{document}